\documentclass{article}

\usepackage[nonatbib,final]{neurips_2020}

\usepackage[utf8]{inputenc} 
\usepackage[T1]{fontenc}    
\usepackage{hyperref}       
\usepackage{url}            
\usepackage{booktabs}       
\usepackage{amsfonts}       
\usepackage{nicefrac}       
\usepackage{microtype}      

\usepackage{amsmath}
\usepackage{common}
\usepackage{graphicx}
\usepackage{todonotes}
\usepackage{float}
\usepackage{physics}
\usepackage{amsmath}
\usepackage{algorithm}
\usepackage[noend]{algpseudocode}
\usepackage{bm}
\usepackage{subcaption}    
\usepackage{cleveref}
\usepackage{natbib}
\usepackage{wrapfig}
\usepackage{fixme}

\title{Short-Term Memory Optimization in Recurrent Neural Networks by Autoencoder-based Initialization}

\author{%
  Antonio Carta \\
  Department of Computer Science\\
  University of Pisa\\
  \texttt{antonio.carta@di.unipi.it} \\
  \And
  Alessandro Sperduti \\
  Department of Mathematics \\
  University of Padova \\
  \texttt{alessandro.sperduti@unipd.it} \\
  \And
  Davide Bacciu \\
  Department of Computer Science \\
  University of Pisa \\
  \texttt{bacciu@di.unipi.it} \\
}

\begin{document}

\maketitle

\begin{abstract}
  Training RNNs to learn long-term dependencies is difficult due to vanishing gradients. We explore an alternative solution based on explicit memorization using linear autoencoders for sequences, which allows to maximize the short-term memory and that can be solved with a closed-form solution without backpropagation. We introduce an initialization schema that pretrains the weights of a recurrent neural network to approximate the linear autoencoder of the input sequences and we show how such pretraining can better support solving hard classification tasks with long sequences. We test our approach on sequential and permuted MNIST. We show that the proposed approach achieves a much lower reconstruction error for long sequences and a better gradient propagation during the finetuning phase. 
\end{abstract}    

\section{Introduction}

Recurrent neural networks are difficult to train due to vanishing gradient problems~\cite{hochreiter1998vanishing,frasconi1994learning} that limit the ability to learn long-term dependencies. Theoretically, long-term dependencies can be easily learned by starting from an hidden state representation  that encodes the entire sequence, giving immediate access to the past history through the hidden state vector. Learning such a representation requires maximizing the \emph{short-term memory (STM) capacity} of the RNN~\cite{jaeger2004esn}. Unfortunately, this objective is hard to optimize by backpropagation (BP) due to the 
vanishing gradient problem.

In this paper, we address the problem of maximizing the STM by avoiding BP using linear autoencoders, which can be easily optimized with a closed-form solution~\cite{preatrain_la_Pasa2014}. We exploit linear autoencoders as a means to initialize RNNs to approximate the linear autoencoder for the input sequences. To reduce the approximation error of such an initialization, caused by the RNN nonlinearity, we propose to use a linear recurrence, removing the nonlinearity in the recurrent update. We show how a recently proposed RNN architecture, the Linear Memory Network (LMN)~\cite{cartaEncodingbasedMemoryModules2020,lmn_carta_icann19}, can be exploited to this end by combining linear and nonlinear units without reducing the expressiveness of the model. We test our approach on a classic benchmark for RNNs, showing that it is competitive with state-of-the-art architectures. Furthermore, we show that the proposed autoencoder initialization maximizes the STM compared to a random initialization.

\section{Optimizing the Short-Term Memory}\label{sec:laes}
In this section, we show how to optimize the short-term memory without BP. The STM capacity measures the ability of a recurrent model to remember and reconstruct past elements of the input sequence~\cite{jaeger2004esn}. We build on a recurrent encoder model $\vh^t = enc(\vx^t, \vh^{t-1})$, and a corresponding decoder $\tilde{\vx}^{t- k} = dec^k (\vh^t)$, which may be a feedforward model trained to reconstruct $\vx^{t-k}$ or a recurrent model applied for $k$ timesteps. We measure the STM with the reconstruction error $E(\vx) = \sum_{k=0}^{t-1} (dec^{i}(\vh^t) - \vx^{t - i})^2$. We can maximize the STM of the encoder by minimizing $E(\vx)$. This objective facilitates learning longer dependencies by allowing to directly reconstruct past elements. Unfortunately, optimizing the STM by BP is difficult due to the vanishing gradient. Furthermore, autoencoding requires to reconstruct the entire input, which means that the maximum dependency length depends on the maximum length of the input sequences. See the additional material (Appendix B) for an example of MNIST images reconstruction with different recurrent models. The reconstructions clearly show that models trained by backpropagation fail to solve autoencoding problems. 

A \emph{linear autoencoder for sequences (LAES)}~\cite{DBLP:conf/icann/Sperduti06,ecml_Sperduti07} is a linear recurrent model composed of a linear dynamical system (encoder) and a linear decoder. The autoencoder takes as input a sequence of vectors $\vx^1, \hdots, \vx^T$ and computes an internal state vector $\vm^t$ as follows:
\begin{align*}
    \vm^t = \mA \vx^t + \mB \vm^{t-1}  \\
    \begin{bmatrix} \tilde{\vx}^t \\ \tilde{\vm}^{t-1} \end{bmatrix} = \mC\vm^t,
\end{align*}
where matrices $\mA$ and $\mB$ are the encoder parameters, $\mC$ is the decoding matrix and $\tilde{\vx}^t$ and $\tilde{\vm}^{t-1}$ are the reconstructions of the current input and the previous memory state. ~\citet{preatrain_la_Pasa2014} provide a closed-form solution for the optimal linear autoencoder. The corresponding decoder parameters $\mC$ can be reconstructed from the encoder as $\mC=\begin{bmatrix} \mA^\top \\ \mB^\top \end{bmatrix}$.

\section{Linear Memory Network}
Using the LAES introduced in Section~\ref{sec:laes}, we can maximize the STM capacity of recurrent neural networks while avoiding the BP through time. However, such a model is limited to a linear encoding of the sequence. Recurrent neural networks have nonlinear recurrent updates that cannot be used to model a LAES. To solve this issue, the Linear Memory Network (LMN)~\cite{lmn_carta_icann19} combines a LAES within a nonlinear recurrent architecture. The LMN computes a hidden state \(\vh^t\) and a separate memory state \(\vm^t\). The hidden state \(\vh^t\) is computed with a nonlinear transformation of the input and previous memory state, while the memory \(\vm^t\) is updated with a linear recurrence using the current hidden state and the previous memory as follows:
\begin{align*}
    \vh^t &= \sigma(\mW_{xh} \vx^t + \mW_{mh} \vm^{t-1})   \\
    \vm^t &= \mW_{hm} \vh^t + \mW_{mm} \vm^{t-1},
\end{align*}
where \(\mW_{xh}, \mW_{mh}, \mW_{hm}, \mW_{mm}\) are the model parameters and \(\sigma\) is a non-linear activation function (\(\tanh\) in our experiments).

Notice that the linearity of the memory update function does not limit the model's expressiveness. In fact, given an RNN with parameters \(\mV\) and \(\mU\) such that \(\vh_{rnn}^t = \sigma(\mV \vx^t + \mU \vh_{rnn}^{t-1})\), we can define an equivalent LMN such that \(\vm^t = \vh_{rnn}^t \forall t\) by setting the parameters as \(\mW_{xh} = \mV, \mW_{mh} = \mU, \mW_{hm} = \mI, \mW_{mm} = 0\).

\section{Initialization with a Linear Autoencoder for Sequences}\label{sec:init}
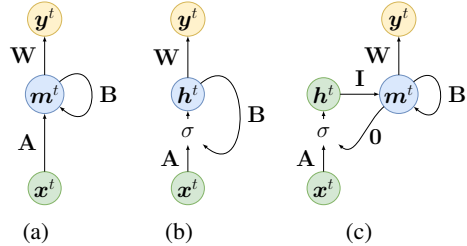
\begin{wrapfigure}{R}{0.5\textwidth}
    \begin{subfigure}{.13\columnwidth}
      \centering
      \scalebox{0.5}{\LARGE{\definecolor{blue_light}{RGB}{218,232,252}
\definecolor{blue_dark}{RGB}{108,142,191}
\definecolor{green_light}{RGB}{213,232,212}
\definecolor{green_dark}{RGB}{130,179,102}
\definecolor{yellow_light}{RGB}{255,242,204}
\definecolor{yellow_dark}{RGB}{214,182,86}

\begin{tikzpicture}
    \tikzstyle{sty_memory} = [draw=blue_dark,fill=blue_light,circle,outer sep=0,inner sep=1,minimum size=25]
    \tikzstyle{sty_functional} = [draw=green_dark,fill=green_light,circle,outer sep=0,inner sep=1,minimum size=25]
    \tikzstyle{sty_out} = [draw=yellow_dark,fill=yellow_light,circle,outer sep=0,inner sep=1,minimum size=25]
    \tikzstyle{myedgestyle} = [-latex]
    \tikzstyle{sty_module} = [rounded corners,fill opacity=0.4]

    
    \node[sty_functional] (x) at (0, -1.5) {$\vx^t$};
    \node[sty_memory] (h) at (0, 1) {$\vm^t$};
    \node[sty_out] (y) at (0, 3) {$\vy^t$};

    \draw [myedgestyle] (x) edge node[left] {$\mA$} (h);
    \draw[myedgestyle] (h) to [out=40,in=-40,loop,looseness=5.8] node[right] {$\mB$} (h);
    \draw [myedgestyle] (h) edge node[left] {$\mW$} (y);

\end{tikzpicture}
}}
      \caption{}\label{fig:lin-rnn}
    \end{subfigure}
    \begin{subfigure}{.13\columnwidth}
      \centering
      \scalebox{0.5}{\LARGE{\definecolor{blue_light}{RGB}{218,232,252}
\definecolor{blue_dark}{RGB}{108,142,191}
\definecolor{green_light}{RGB}{213,232,212}
\definecolor{green_dark}{RGB}{130,179,102}
\definecolor{yellow_light}{RGB}{255,242,204}
\definecolor{yellow_dark}{RGB}{214,182,86}

\begin{tikzpicture}
    \tikzstyle{sty_memory} = [draw=blue_dark,fill=blue_light,circle,outer sep=0,inner sep=1,minimum size=25]
    \tikzstyle{sty_functional} = [draw=green_dark,fill=green_light,circle,outer sep=0,inner sep=1,minimum size=25]
    \tikzstyle{sty_out} = [draw=yellow_dark,fill=yellow_light,circle,outer sep=0,inner sep=1,minimum size=25]
    \tikzstyle{myedgestyle} = [-latex]
    \tikzstyle{sty_module} = [rounded corners,fill opacity=0.4]

    
    \node[sty_functional] (x) at (0, -1.5) {$\vx^t$};
    \node[sty_memory] (h) at (0, 1) {$\vh^t$};
    \node[] (sigma) at (0, 0) {$\sigma$};
    \node[sty_out] (y) at (0, 3) {$\vy^t$};

    \draw [myedgestyle] (x) edge node[left] {$\mA$} (sigma);
    \draw[myedgestyle] (h) to [out=40,in=-40,loop,looseness=2.8] node[right] {$\mB$} (sigma);
    \draw [myedgestyle] (sigma) edge (h);
    \draw [myedgestyle] (h) edge node[left] {$\mW$} (y);

\end{tikzpicture}
}}
      \caption{}\label{fig:rnn-laes}
    \end{subfigure}
    \begin{subfigure}{.2\columnwidth}
        \centering
        \scalebox{0.5}{\LARGE{\definecolor{blue_light}{RGB}{218,232,252}
\definecolor{blue_dark}{RGB}{108,142,191}
\definecolor{green_light}{RGB}{213,232,212}
\definecolor{green_dark}{RGB}{130,179,102}
\definecolor{yellow_light}{RGB}{255,242,204}
\definecolor{yellow_dark}{RGB}{214,182,86}

\begin{tikzpicture}
    \tikzstyle{sty_memory} = [draw=blue_dark,fill=blue_light,circle,outer sep=0,inner sep=1,minimum size=25]
    \tikzstyle{sty_functional} = [draw=green_dark,fill=green_light,circle,outer sep=0,inner sep=1,minimum size=25]
    \tikzstyle{sty_out} = [draw=yellow_dark,fill=yellow_light,circle,outer sep=0,inner sep=1,minimum size=25]
    \tikzstyle{myedgestyle} = [-latex]
    \tikzstyle{sty_module} = [rounded corners,fill opacity=0.4]

    
    \node[sty_functional] (x) at (0, -1.5) {$\vx^t$};
    \node[sty_functional] (h) at (0, 1) {$\vh^t$};
    \node[] (sigma) at (0, 0) {$\sigma$};
    \node[sty_memory] (m) at (2, 1) {$\vm^t$};
    \node[sty_out] (y) at (2, 3) {$\vy^t$};

    \draw [myedgestyle] (x) edge node[left] {$\mA$} (sigma);
    \draw[myedgestyle] (h) to node[above] {$\mI$} (m);
    \draw[myedgestyle] (m) to [out=220,in=-40] node[right] {$\mathbf{0}$} (sigma);
    \draw[myedgestyle] (m) to [out=40,in=-40,loop,looseness=4.8] node[right] {$\mB$} (m);
    \draw [myedgestyle] (sigma) edge (h);
    \draw [myedgestyle] (m) edge node[left] {$\mW$} (y);

\end{tikzpicture}
}}
        \caption{}\label{fig:lmn-laes}
    \end{subfigure}
    \caption{Linear RNN (Fig. \ref{fig:lin-rnn}). Initialization schemes for RNN (Fig. \ref{fig:rnn-laes}), and LMN (Fig. \ref{fig:lmn-laes}). Notice how the nonlinearity only affects the input in the LMN with LAES initialization. Blue nodes highlight the LAES memory state.}\label{fig:fig}
\end{wrapfigure}

We can use the LAES to initialize recurrent models to encode the input sequences into a minimal hidden state, increasing their short-term memory compared to random initializations. We define a linear recurrent model which uses the autoencoder to encode the input sequences within a single vector, i.e. the memory state of the autoencoder. To predict the desired target, we can train a linear readout that takes as input the states of the autoencoder as follows:
\begin{align}
    \vm^t = \mA \vx^t + \mB \vm^{t-1}  \label{eq:lin-ae-rnn} \\ 
    \vy^t = \mW_o \vm^t,               \label{eq:lin-ro-rnn} 
\end{align}
where $\mA$ and $\mB$ are the parameters of the autoencoder trained to reconstruct the input sequences and $\mW_o$ are the parameters of a linear model trained to predict the target output from $\vm^t$. Figure~\ref{fig:lin-rnn} shows a schematic view of the architecture.

~\citet{preatrain_la_Pasa2014} propose to initialize an RNN with the linear RNN defined in Eq.~\ref{eq:lin-ae-rnn} and \ref{eq:lin-ro-rnn} as:
\begin{align*}
    \vh^t = \tanh(\mA \vx^t + \mB \vh^{t-1})  \\
    \vy^t = \mW_o \vh^t.
\end{align*}
Notice that the initialization above is an approximation of the original LAES due to the presence of the  nonlinear activation function. The quality of such an approximation depends on the values of the hidden activations $\vm^t$. For small values close to zero, the $tanh$ activation function is approximately linear, and therefore the approximation error is negligible. However, larger hidden activation values saturate the $\tanh$ function, incuring in large approximation errors. In practice, we find that the correspondence between the linear autoencoder and the initialized RNN degrades quickly due to the accumulation of errors through time. To solve this problem, we put forward the use of a LMN, which provides a linear recurrence, in place of the RNN above. The LMN model is initialized as follows:
\begin{align*}
    \vh^t = \sigma(\mA \vx^t + \textbf{0} \vm^{t-1})   \\
    \vm^t = \textbf{I} \vh^t + \mB \vm^{t-1}   \\
    \vy^t = \mW_o \vm^t,
\end{align*}
where the parameters are $\mW_{xh}=\mA, \mW_{mh}=0$, $\mW_{hm} = \mI$, $\mW_{mm}={\mB}$. Such an initialization is an approximation of the linear RNN, but since the nonlinearity only affects the input transformation $\mA\vx^t$ and not the recurrence, the approximation is closer to the original linear RNN. In practice, we did not find any difference between the linear RNN and LMN accuracy after the initialization. Figures \ref{fig:rnn-laes} and \ref{fig:lmn-laes} show a schematic view of the architectures.

\section{Experiments} \label{sec:experiments}
We evaluate the performance of the autoencoder initialization and its effect on gradient propagation. Furthermore, we want to verify if finetuning of the recurrent units by BP is necessary. We compare the following models, along with baselines from the literature, the RNN and LMN, both with orthogonal or LAES initialization:
\begin{description}
    \item[LAES-Linear] A LAES with a linear classifier trained by pseudoinverse. The model does not use BP.
    \item[LAES-SVM] A LAES with an SVM classifier. The model does not use BP.
    \item[LAES-FF] A LAES with a feedforward classifier. The model uses truncated BP, propagating the gradient only through the feedforward units, leaving the recurrent connections fixed. 
\end{description}
Please check the additional material for details about the chosen baselines and the experimental setup. We test the proposed model on sequential and permuted MNIST~\cite{irnn_seqmnist_le2015}, two popular benchmarks with long sequences ($784$ elements). We also measure the effect of the LAES initialization on the gradient propagation and sequence reconstruction. The source code for the experiments is available online\footnote{\url{https://github.com/AntonioCarta/rnn_autoencoding_neurips2020}}.

\subsection{Sequential and Permuted MNIST}
Table \ref{tbl:res_mnist} shows the results.The best results are achieved by the AntisymmetricRNN~\cite{antisymmetric_rnn_Chang2019} for sequential MNIST and the LMN with LAES initialization for permuted MNIST. The LAES initialization consistently improves over the orthogonal initialization for RNN and LMN. 
\begin{wraptable}{r}{0.4\textwidth}
        \centering
        \caption{Test accuracy on sequential MNIST and permuted MNIST. Experiments denoted with $\dag$, $\diamond$, $\ddagger$, $\bullet$ are taken from~\protect\cite{wisdom_full_urnn_nips2016},~\protect\cite{jingli_eunn_icml17},~\protect\cite{jose_kronecker_rnn_icml18}, and~\protect\cite{antisymmetric_rnn_Chang2019} respectively.} \label{tbl:res_mnist}
        \resizebox{.4\textwidth}{!}{
            \begin{tabular}{llll}
            \toprule
                                    &       & \multicolumn{2}{c}{MNIST} \\
                                    & $n_h$ & Seq. & Perm. \\
            \midrule
            FC uRNN$^\dag$          & 512   & 96.9 & 94.1 \\
            EURNN$^\diamond$        & 1024  & -    & 93.7 \\
            KRU$^\ddagger$          & 512   & 96.4 & 94.5 \\
            LSTM$\ddagger$          & 128   & 97.8 & 91.3 \\ 
            ASRNN\(^\bullet\)       & 128   & 98.0 & 95.8 \\ 
            ASRNN(gated)\(^\bullet\) & 128  & \textbf{98.8} & 93.1 \\
            \midrule
            LAES-Linear             & 128   & 86.6 & 84.2 \\
            LAES-SVM                & 128   & 94.7 & 86.8 \\
            LAES-FF                 & 128   & 97.9 & 92.3 \\
            \midrule
            RNN-ortho               & 128   & 77.2 & 84.9 \\
            RNN-laes                & 128   & 97.0 & 91.8 \\
            \midrule
            LMN-ortho               & 128   & 95.3 & 94.6 \\
            LMN-laes                & 128   & 98.5 & \textbf{96.0} \\
            \bottomrule
        \end{tabular}}  
\end{wraptable} 
Overall, the LMN with LAES initialization provides the most consistent results on average on the two datasets ($97.3$ against $96.9$ in ASRNN). We notice that the LAES initialization is expressive even without finetuning since the LAES-FF obtains competitive results, especially on sequential MNIST, which is easier to encode. However, finetuning with BP remains a necessary step to increase the final performance.
\begin{wrapfigure}{R}{0.52\textwidth}
    \centering
    \begin{subfigure}{0.25\columnwidth}
        \centering
        \includegraphics[width=\textwidth]{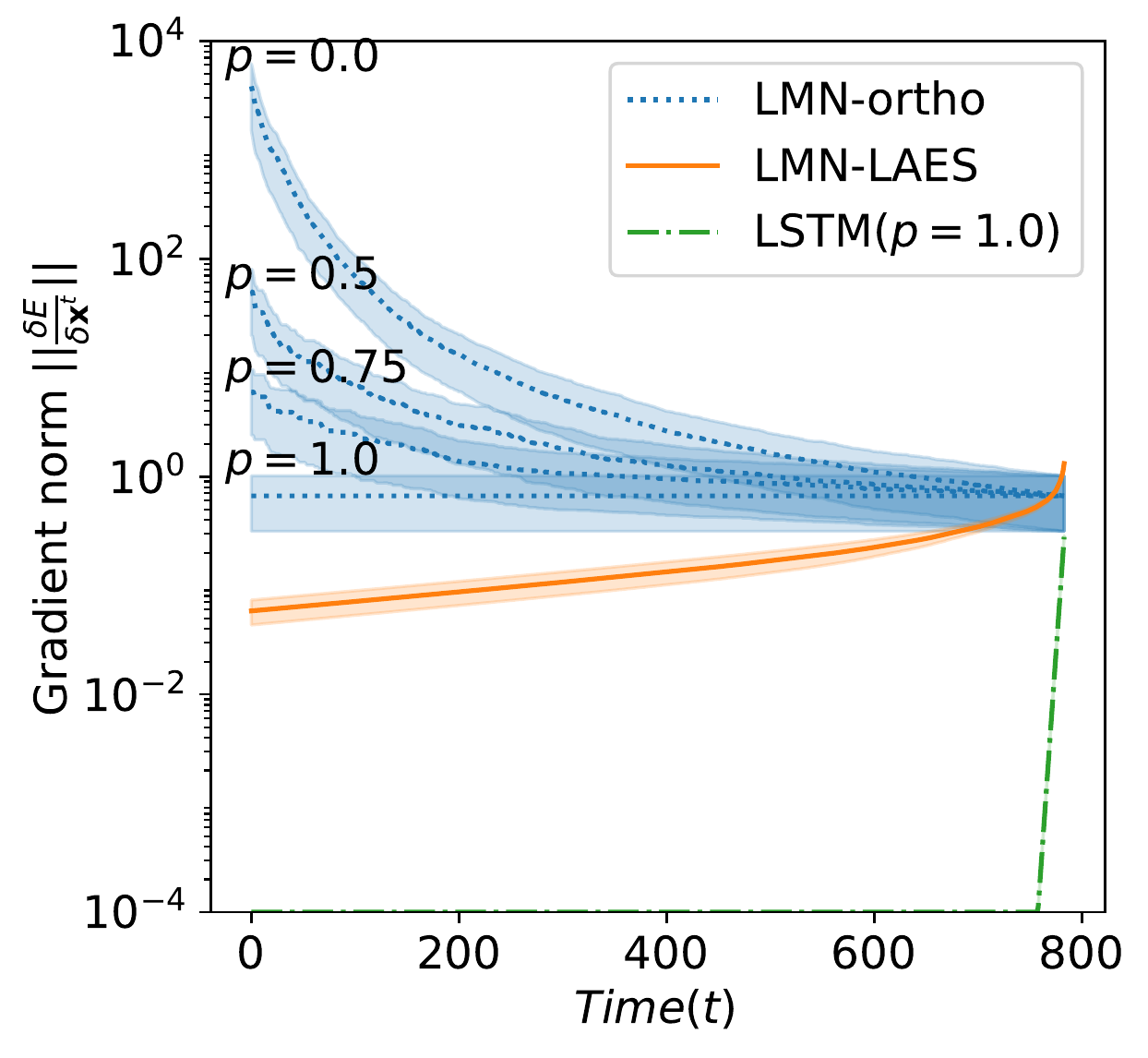}
        \caption{}\label{fig:grad_init}
    \end{subfigure}
    \begin{subfigure}{0.25\columnwidth}
        \centering
        \includegraphics[width=\textwidth]{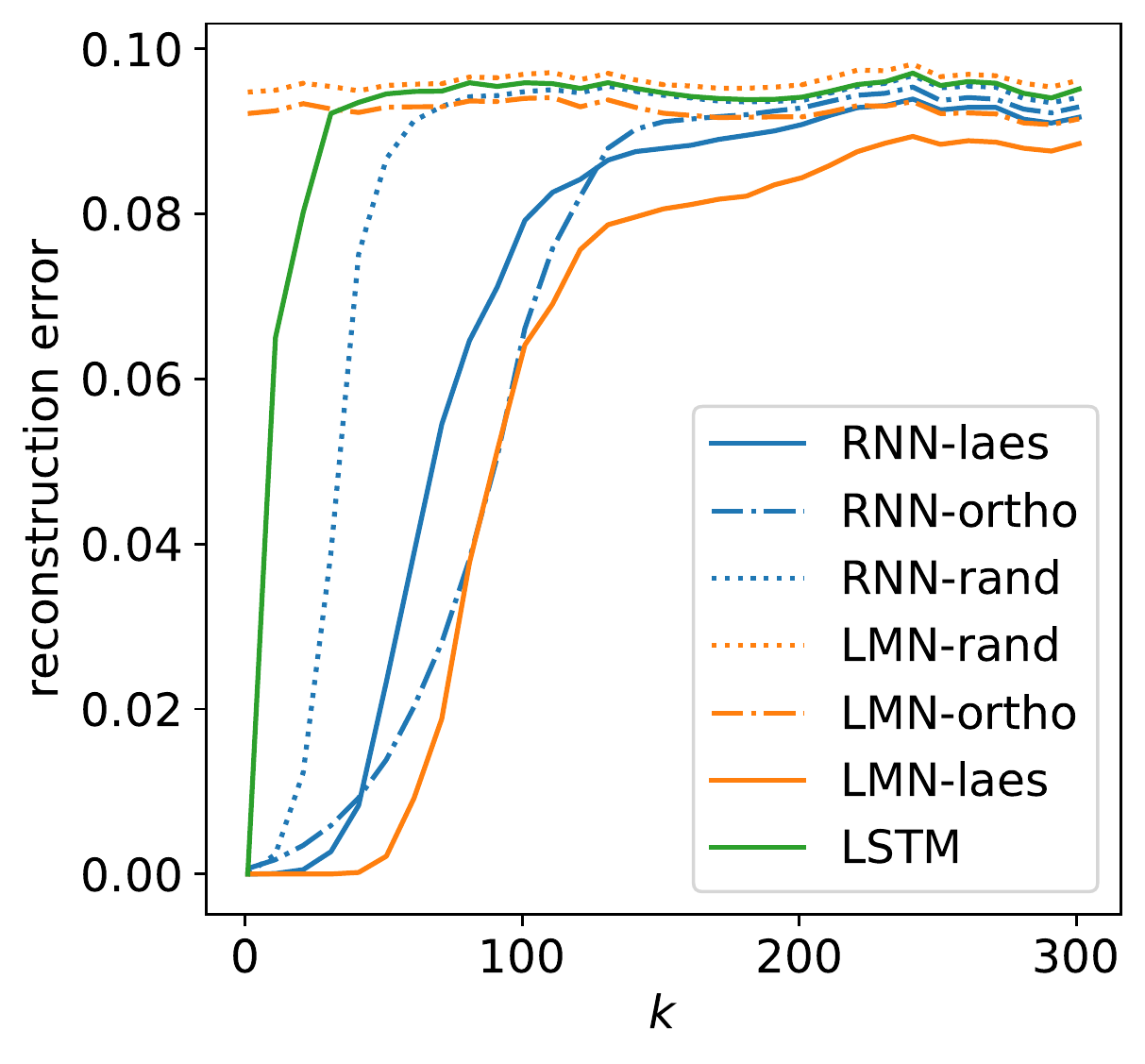}
        \caption{}\label{fig:reco_err}
    \end{subfigure}

    \caption{Gradient propagation through time (Fig. \ref{fig:grad_init}) and reconstruction error (Fig. \ref{fig:reco_err}).}\label{fig:my_label}
\end{wrapfigure}

Recurrent models such as the Elman RNN and LSTM suffer from vanishing gradients. Figure~\ref{fig:grad_init} shows the gradient propagation through time for the LMN with orthogonal and LAES initialization, and the LSTM. For the orthogonal LMN, we plot the gradient for different values of the probability used to truncate the gradient \(p\)~\cite{h_detach_Kanuparthi2018}. The error signal \(\pdv{E}{\vh^T}\) at the last timestep \(T=784\) is backpropagated to the start of the sequence at \(t=0\), obtaining \(\pdv{E}{\vh^0}\). The norm of the gradient shows the evolution of the gradient through time. Learning long-term dependencies is easier if the gradient can be propagated through large intervals without a large vanishing or exploding effect. 

The LMN with orthogonal initialization and full gradient (\(p=0\)) shows exploding gradients, while full truncation (\(p=1\)) shows a constant propagation (as can be shown analytically). The effect of the exploding gradient can be gradually mitigated by increasing \(p\) from $0$ to $1$. The LSTM suffers from the vanishing gradient and annihilates the gradients after less than \(50\) steps, rendering it useless to learn long-term dependencies. The LMN with LAES initialization has a vanishing gradient. However, the slope of its curve is small and after \(t=784\) backpropagation steps the norm of the gradient is still large enough to allow learning long-term dependencies.

Figure \ref{fig:reco_err} shows the reconstruction error for different architectures and initializations. A linear model is trained to reconstruct the hidden activations \(\vx^{t-k}\) given \(\vh^t\), with a separate model trained for each \(k \in \{1, 300\}\). We plot the average reconstruction error for each configuration. The LMN with LAES initialization obtains the best performance for both short-term and long-term reconstruction. Classic models such as the LSTM and orthogonal models obtain worse reconstructions.

\section{Conclusion}

In this work, we propose the maximization of the STM capacity as an alternative objective to encourage learning long-term dependencies with recurrent neural networks. Maximizing the STM is a difficult problem due to the presence of long-term dependencies. We show that RNN models trained with BP fail to encode simple but long sequences. To address this issue, we propose to initialize recurrent models with a linear autoencoder for sequences. The LAES can be trained without BP and efficiently encodes long sequences, unlike RNNs trained with BP. As a future work, we plan to improve the proposed method to completely remove the BP step through the recurrent units.

\bibliography{biblio}

\begin{thebibliography}{}

\bibitem[\protect\citeauthoryear{Arjovsky \bgroup \em et al.\egroup
  }{2015}]{unitary_Arjovsky2015}
M.~Arjovsky, A.~Shah, and Y.~Bengio.
\newblock Unitary evolution recurrent neural networks.
\newblock In {\em ICML}, 2015.

\bibitem[\protect\citeauthoryear{Arpit \bgroup \em et al.\egroup
  }{2018}]{h_detach_Kanuparthi2018}
D.~Arpit, B.~Kanuparthi, G.~Kerg, Nan~R. Ke, I.~Mitliagkas, and Y.~Bengio.
\newblock h-detach: Modifying the lstm gradient towards better optimization.
\newblock {\em ArXiv}, abs/1810.03023, 2018.

\bibitem[\protect\citeauthoryear{Bacciu \bgroup \em et al.\egroup
  }{2019}]{lmn_carta_icann19}
D.~Bacciu, A.~Carta, and A.~Sperduti.
\newblock {Linear Memory Networks}.
\newblock In {\em ICANN}, 2019.

\bibitem[\protect\citeauthoryear{Bengio \bgroup \em et al.\egroup
  }{1994}]{frasconi1994learning}
Yoshua Bengio, Patrice Simard, and Paolo Frasconi.
\newblock {Learning long-term dependencies with gradient descent is difficult}.
\newblock {\em IEEE transactions on neural networks}, 5(2):157--166, 1994.

\bibitem[\protect\citeauthoryear{Carta \bgroup \em et al.\egroup
  }{2020a}]{cartaEncodingbasedMemoryModules2020}
Antonio Carta, Alessandro Sperduti, and Davide Bacciu.
\newblock Encoding-based {{Memory Modules}} for {{Recurrent Neural Networks}}.
\newblock {\em arXiv:2001.11771 [cs, stat]}, January 2020.

\bibitem[\protect\citeauthoryear{Carta \bgroup \em et al.\egroup
  }{2020b}]{cartaIncrementalTrainingRecurrent2020}
Antonio Carta, Alessandro Sperduti, and Davide Bacciu.
\newblock Incremental {{Training}} of a {{Recurrent Neural Network Exploiting}}
  a {{Multi}}-{{Scale Dynamic Memory}}.
\newblock {\em arXiv:2006.16800 [cs, stat]}, June 2020.

\bibitem[\protect\citeauthoryear{Casado and
  Mart{\'i}nez-Rubio}{2019}]{Lezcano-Casado2019}
M.~Lezcano Casado and D.~Mart{\'i}nez-Rubio.
\newblock Cheap orthogonal constraints in neural networks: A simple
  parametrization of the orthogonal and unitary group.
\newblock In {\em ICML}, 2019.

\bibitem[\protect\citeauthoryear{Chang \bgroup \em et al.\egroup
  }{2019}]{antisymmetric_rnn_Chang2019}
B.~Chang, M.~Chen, E.~Haber, and Ed~H. Chi.
\newblock Antisymmetricrnn: A dynamical system view on recurrent neural
  networks.
\newblock {\em ArXiv}, abs/1902.09689, 2019.

\bibitem[\protect\citeauthoryear{Ganguli \bgroup \em et al.\egroup
  }{2008}]{Ganguli2008}
S.~Ganguli, D.~Huh, and H.~Sompolinsky.
\newblock {Memory Traces in Dynamical Systems - Supplementary Material
  Contents}.
\newblock {\em Proceedings of the National Academy of Sciences}, 2008.

\bibitem[\protect\citeauthoryear{Hochreiter and
  Schmidhuber}{1997}]{hochreiter_lstm97}
S.; Hochreiter and J.; Schmidhuber.
\newblock {Long Short-Term Memory}.
\newblock {\em Neural Computation}, 9(8):1--32, 1997.

\bibitem[\protect\citeauthoryear{Hochreiter}{1998}]{hochreiter1998vanishing}
Sepp Hochreiter.
\newblock {The vanishing gradient problem during learning recurrent neural nets
  and problem solutions}.
\newblock {\em International Journal of Uncertainty, Fuzziness and
  Knowledge-Based Systems}, 6(02):107--116, 1998.

\bibitem[\protect\citeauthoryear{Jaeger and Haas}{2004}]{jaeger2004esn}
H.~Jaeger and H.~Haas.
\newblock {Harnessing nonlinearity: Predicting chaotic systems and saving
  energy in wireless communication}.
\newblock {\em science}, 304(5667), 2004.

\bibitem[\protect\citeauthoryear{Jing \bgroup \em et al.\egroup
  }{2016}]{jingli_eunn_icml17}
L.~Jing, Y.~Shen, T.~Dub{\v{c}}ek, J.~Peurifoy, S.~Skirlo, Y.~LeCun,
  M.~Tegmark, and M.~Solja{\v{c}}i{\'{c}}.
\newblock {Tunable Efficient Unitary Neural Networks (EUNN) and their
  application to RNNs}.
\newblock In {\em ICML}, 2016.

\bibitem[\protect\citeauthoryear{Jose \bgroup \em et al.\egroup
  }{2018}]{jose_kronecker_rnn_icml18}
C.~Jose, M.~Ciss{\'{e}}, and F.~Fleuret.
\newblock {Kronecker Recurrent Units}.
\newblock In {\em ICML}, may 2018.

\bibitem[\protect\citeauthoryear{Kingma and Ba}{2014}]{adam_kingma2014}
D.~P. Kingma and J.~Ba.
\newblock {Adam: A Method for Stochastic Optimization}.
\newblock {\em arXiv preprint arXiv:1412.6980}, 2014.

\bibitem[\protect\citeauthoryear{Krueger and Memisevic}{2015}]{Krueger2015}
D.~Krueger and R.~Memisevic.
\newblock Regularizing rnns by stabilizing activations.
\newblock {\em CoRR}, abs/1511.08400, 2015.

\bibitem[\protect\citeauthoryear{Le \bgroup \em et al.\egroup
  }{2015}]{irnn_seqmnist_le2015}
Quoc~V Le, N.~Jaitly, and G.~E Hinton.
\newblock {A simple way to initialize recurrent networks of rectified linear
  units}.
\newblock {\em arXiv preprint arXiv:1504.00941}, 2015.

\bibitem[\protect\citeauthoryear{LeCun}{1998}]{lecun1998mnist}
Yann LeCun.
\newblock {The MNIST database of handwritten digits}.
\newblock {\em http://yann. lecun. com/exdb/mnist/}, 1998.

\bibitem[\protect\citeauthoryear{Mhammedi \bgroup \em et al.\egroup
  }{2017}]{householder_ornn_icml17}
Z.~Mhammedi, A.~Hellicar, A.~Rahman, and J.~Bailey.
\newblock {Efficient Orthogonal Parametrisation of Recurrent Neural Networks
  Using Householder Reflections}.
\newblock In {\em ICML}, 2017.

\bibitem[\protect\citeauthoryear{Pasa and
  Sperduti}{2014}]{preatrain_la_Pasa2014}
L.~Pasa and A.~Sperduti.
\newblock {Pre-training of Recurrent Neural Networks via Linear Autoencoders}.
\newblock {\em NIPS}, 2014.

\bibitem[\protect\citeauthoryear{Sperduti}{2006}]{DBLP:conf/icann/Sperduti06}
A.~Sperduti.
\newblock {Exact Solutions for Recursive Principal Components Analysis of
  Sequences and Trees}.
\newblock In {\em ICANN}, 2006.

\bibitem[\protect\citeauthoryear{Sperduti}{2007}]{ecml_Sperduti07}
A.~Sperduti.
\newblock Efficient computation of recursive principal component analysis for
  structured input.
\newblock In {\em ECML}, 2007.

\bibitem[\protect\citeauthoryear{Sperduti}{2013}]{sperduti2013linear}
Alessandro Sperduti.
\newblock {Linear autoencoder networks for structured data}.
\newblock In {\em International Workshop on Neural-Symbolic Learning and
  Reasoning}, 2013.

\bibitem[\protect\citeauthoryear{Tino \bgroup \em et al.\egroup
  }{2004}]{markovian_bias_Tino2004}
P.~Tino, M.~Cernansky, and L.~Benuskova.
\newblock {Markovian Architectural Bias of Recurrent Neural Networks}.
\newblock {\em IEEE Transactions on Neural Networks}, 2004.

\bibitem[\protect\citeauthoryear{Tiňo and Rodan}{2013}]{stm_input_tino2013}
P~Tiňo and A~Rodan.
\newblock {Short term memory in input-driven linear dynamical systems}.
\newblock {\em Neurocomputing}, 112:58--63, 2013.

\bibitem[\protect\citeauthoryear{Vorontsov \bgroup \em et al.\egroup
  }{2017}]{vorontsov_ortho_rnn_icml17}
E.~Vorontsov, C.~Trabelsi, S.~Kadoury, and C.~Pal.
\newblock {On orthogonality and learning recurrent networks with long term
  dependencies}.
\newblock In {\em ICML}, 2017.

\bibitem[\protect\citeauthoryear{White \bgroup \em et al.\egroup
  }{2004}]{White2004}
O.~L. White, D.~Lee, and H.~Sompolinsky.
\newblock {Short-term memory in orthogonal neural networks}.
\newblock {\em Physical Review Letters}, 92(14):0--3, 2004.

\bibitem[\protect\citeauthoryear{Wisdom \bgroup \em et al.\egroup
  }{2016}]{wisdom_full_urnn_nips2016}
S.~Wisdom, T.~Powers, J.~R. Hershey, J.~Le Roux, and L.~Atlas.
\newblock {Full-Capacity Unitary Recurrent Neural Networks}.
\newblock In {\em NIPS}, 2016.

\bibitem[\protect\citeauthoryear{Zhang \bgroup \em et al.\egroup
  }{2018}]{zhang_svd_ornn_icml18}
J.~Zhang, Q.~Lei, and I.~Dhillon.
\newblock {Stabilizing Gradients for Deep Neural Networks via Efficient SVD
  Parameterization}.
\newblock In {\em ICML}, 2018.

\end{thebibliography}
\bibliographystyle{named}

\appendix
\section{Related Work}

Orthogonal RNNs solve the vanishing gradient problem by parameterizing the recurrent connections with an orthogonal or unitary matrix~\cite{unitary_Arjovsky2015}.
Some orthogonal models exploit specific parameterizations or factorizations~\cite{householder_ornn_icml17,jose_kronecker_rnn_icml18,jingli_eunn_icml17} to guarantee the orthogonality. Other approaches constrain the parameters with soft or hard orthogonality constraints~\cite{vorontsov_ortho_rnn_icml17,zhang_svd_ornn_icml18,wisdom_full_urnn_nips2016,Lezcano-Casado2019}. ~\cite{vorontsov_ortho_rnn_icml17} have shown that hard orthogonality constraints can hinder training speed and final performance.

Linear autoencoders for sequences can be trained optimally with a closed-form solution~\cite{sperduti2013linear}. They have been used to pretrain RNNs~\cite{preatrain_la_Pasa2014}. The LMN ~\cite{lmn_carta_icann19} is a recurrent neural network with a separate linear recurrence, which can be exploited to memorize long sequences by pretraining it with the LAES~\cite{cartaEncodingbasedMemoryModules2020,cartaIncrementalTrainingRecurrent2020}.

The memorization properties of untrained models are studied in the field of echo state echo networks~\cite{jaeger2004esn}, a recurrent model with untrained recurrent parameters. ~\cite{markovian_bias_Tino2004} showed that untrained RNNs with a random weight initialization have a Markovian bias, which results in the clustering of input with similar suffixes in the hidden state space. ~\cite{stm_input_tino2013,White2004,Ganguli2008} study the short-term memory properties of linear and orthogonal memories.

\section{Sequential and Permuted MNIST experiments}
Sequential and permuted MNIST ~\cite{irnn_seqmnist_le2015} are two standard benchmarks designed to test the ability of recurrent neural networks to learn long-term dependencies. They consist of sequences extracted from the MNIST dataset~\cite{lecun1998mnist} by scanning each image one pixel at a time, in order (sequential MNIST) or with a random fixed permutation (permuted MNIST). We use these datasets to compare the LAES initialization to a random orthogonal initialization, since they provide an ideal setting to test a pure memorization approach like the LAES initialization. All the models have $128$ hidden units and a single hidden layer. Each model has been trained for $100$ epochs with a batch size of $64$.

Each model is trained with Adam~\cite{adam_kingma2014}, with a learning rate in $\{10^{-3}, 10^{-4}, 10^{-5}\}$ chosen by selecting the best model on a separate validation set. A soft orthogonality constraint $\lambda \lVert \mW^\top \mW - \mI \lVert ^ 2$ is added to the cost function as in ~\cite{vorontsov_ortho_rnn_icml17}, with $\lambda$ chosen $\{0, 10^{-5}, 10^{-4}, 10^{-3}, 10^{-2}\}$. We apply the activation regularization proposed in~\cite{Krueger2015}, with \(\alpha \in \{0, 1, 10, 100\}\). To apply the stochastic gradient truncation~\cite{h_detach_Kanuparthi2018} we select the probability of truncation $p \in \{ 0, 0.1, 0.25, 0.5, 1 \}$, which includes the full backpropagation($p=0$) and the exact truncation ($p=1$).

\subsection{Baseline Models}
We compare against orthogonal and unitary RNNs, which are recurrent models with unitary recurrent matrices that guarantee constant gradient propagation\cite{unitary_Arjovsky2015}, such as the FC uRNN\cite{wisdom_full_urnn_nips2016}, EURNN\cite{jingli_eunn_icml17}, KRU\cite{jose_kronecker_rnn_icml18}. We also compare against the LSTM\cite{hochreiter_lstm97}, and ASRNN\cite{antisymmetric_rnn_Chang2019}.

We study different baseline that exploit the LAES to verify if the backpropagation through the hidden units is necessary:
\begin{description}
    \item[LAES-Linear] A LAES with a linear classifier trained by pseudoinverse. The model does not use backpropagation.
    \item[LAES-SVM] A LAES with an SVM classifier. The model does not use backpropagation.
    \item[LAES-FF] A LAES with a feedforward classifier. The model uses truncated backpropagation, propagating the gradient only through the feedforward units, leaving the recurrent connections fixed. 
\end{description}

\section{MNIST Sequence Reconstruction}
\begin{figure}[h!]
    \centering
    \begin{subfigure}[t]{.19\textwidth}
        \centering
        \includegraphics[width=.9\linewidth]{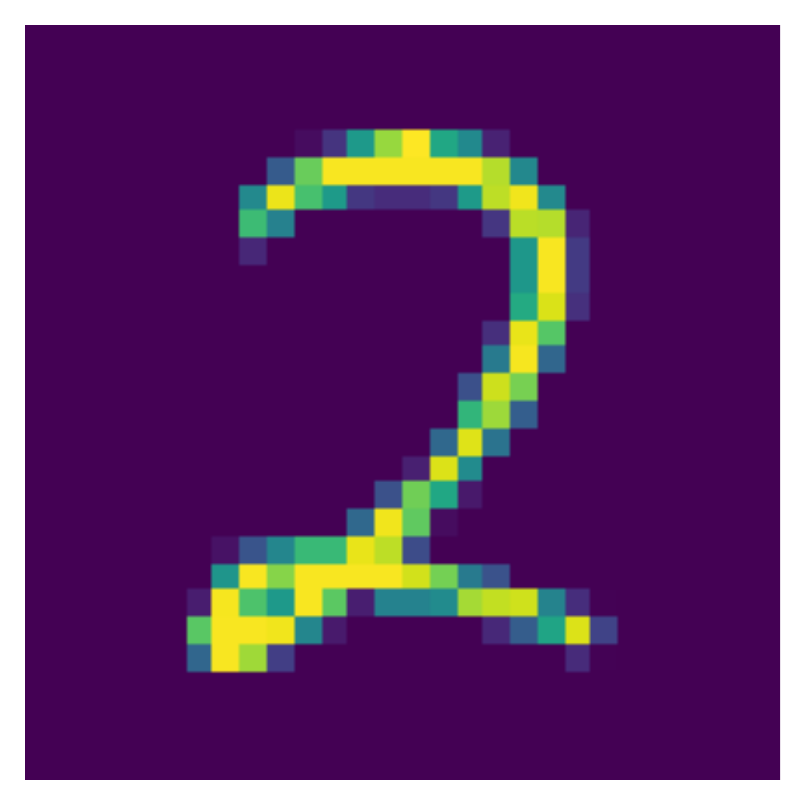}
        \caption{Input image}\label{fig:sgdreco-original}
    \end{subfigure}
    \begin{subfigure}[t]{.19\textwidth}
        \centering
        \includegraphics[width=.9\linewidth]{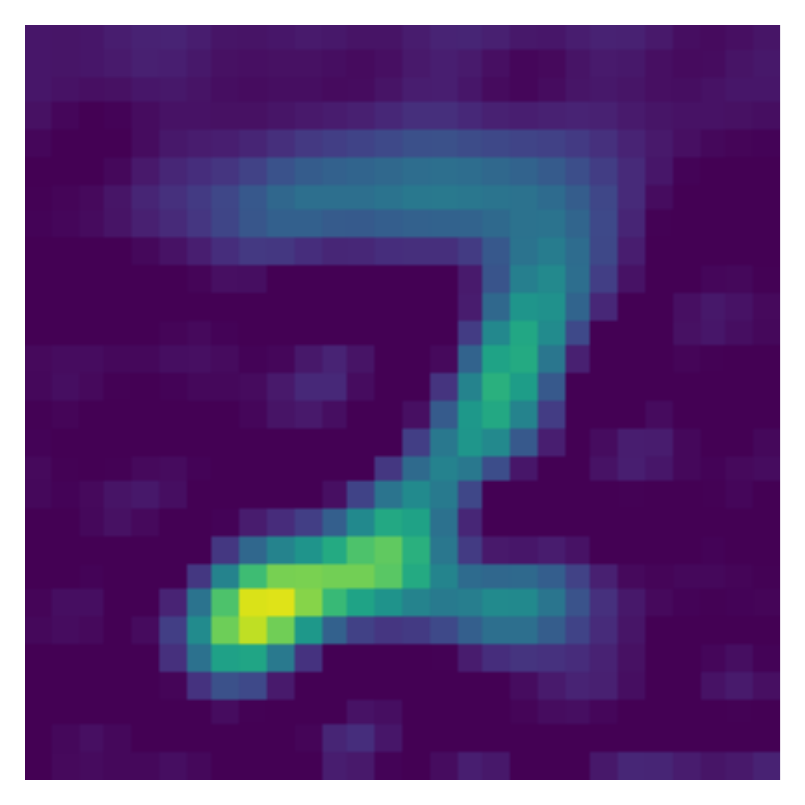}
        \caption{LAES}\label{fig:sgdreco-laes}
    \end{subfigure}
    \begin{subfigure}[t]{.19\textwidth}
        \centering
        \includegraphics[width=.9\linewidth]{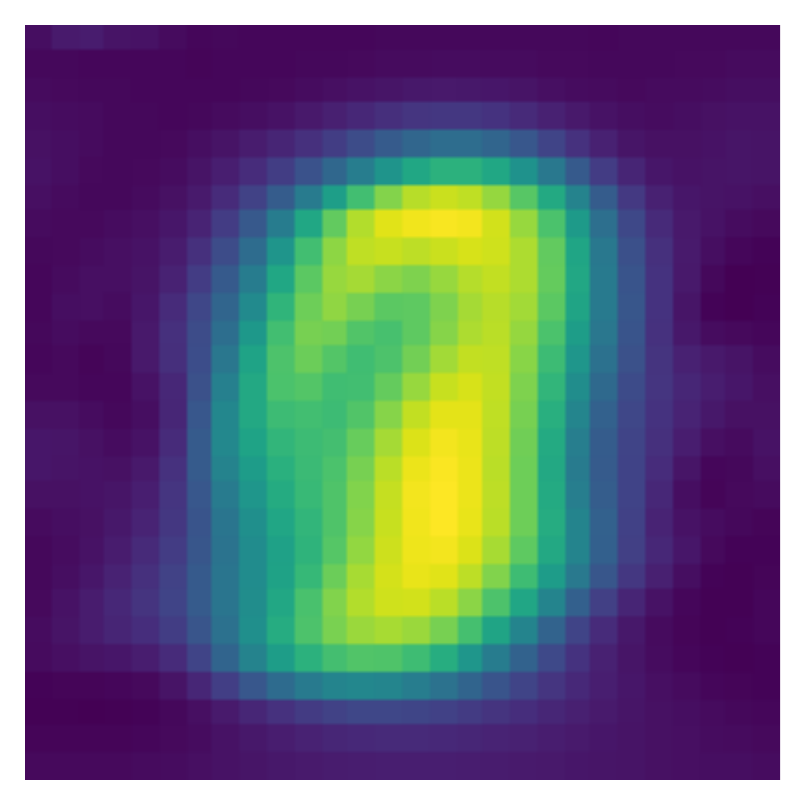}
        \caption{Elman RNN}\label{fig:sgdreco-rnn}
    \end{subfigure}
    \begin{subfigure}[t]{.19\textwidth}
        \centering
        \includegraphics[width=.9\linewidth]{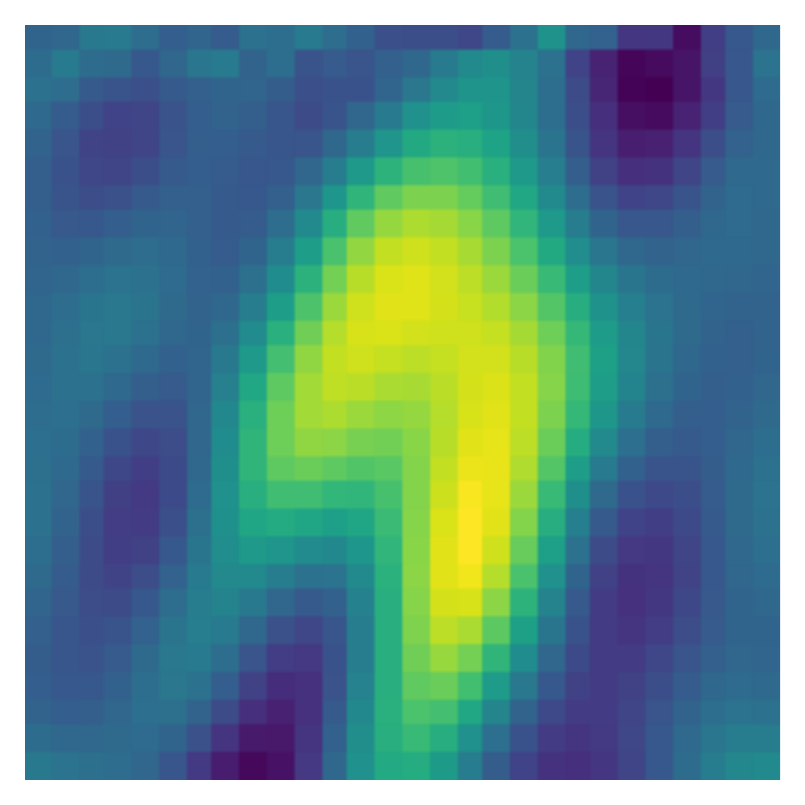}
        \caption{Orthogonal RNN}\label{fig:sgdreco-idrnn}
    \end{subfigure}
    \begin{subfigure}[t]{.19\textwidth}
        \centering
        \includegraphics[width=.9\linewidth]{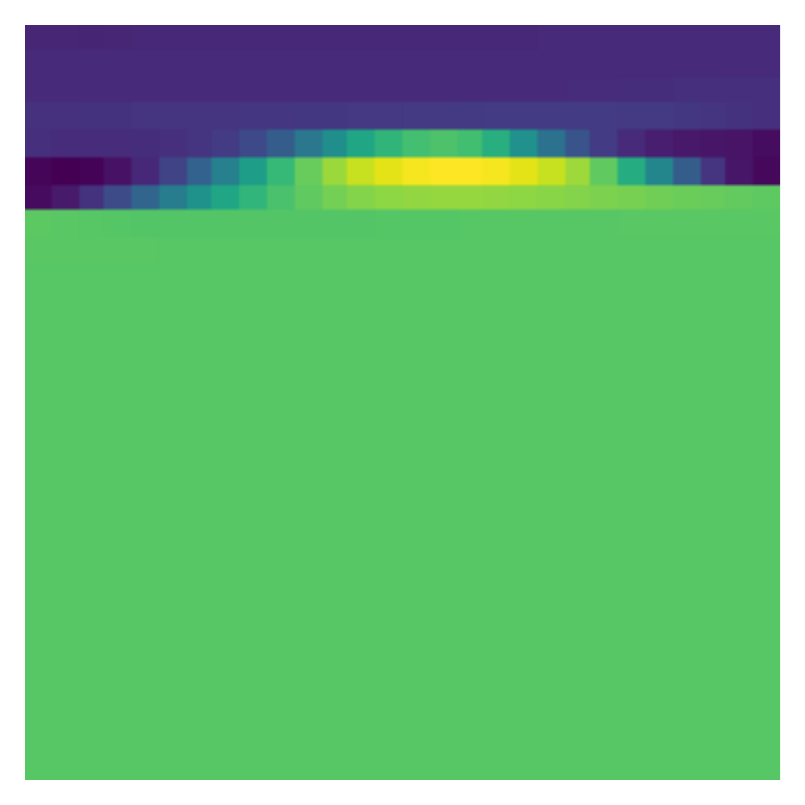}
        \caption{LSTM}\label{fig:sgdreco-lstm}
    \end{subfigure}
    \caption{Sequence reconstructions on Sequential MNIST for recurrent models trained with backpropagation against the LAES.}\label{fig:sgdreco}
\end{figure}

We trained different recurrent models to reconstruct Sequential MNIST samples. Each model consists of a single layer RNN with $100$ hidden units. We compare the Elman RNN, LSTM, RNN with orthogonal initialization, and the LAES. Except for the LAES, each model has been trained for $100$ epochs with backpropagation, using Adam~\cite{adam_kingma2014}, with a learning rate  chosen with a separate validation set in $\{10^{-3},\ 10^{-4}\}$. For the orthogonal RNN we add a soft orthogonality constraint by penalizing the recurrent weight matrix as in \cite{vorontsov_ortho_rnn_icml17}, chosen from the values in $\{0,\ 10^{-1},\ 10^{-2},\ 10^{-3},\ 10^{-4}\}$. Figure \ref{fig:sgdreco} shows a sample reconstruction for each model. The only model which is able to approximately reconstruct the original image is the LAES, which is also the only one not trained with backpropagation.

\end{document}